\pdfoutput=1

\documentclass[11pt]{article}

\usepackage{acl}

\usepackage{times}
\usepackage{latexsym}
\usepackage{bm}
\usepackage{url}
\usepackage{tikz}
\usepackage{booktabs}
\usepackage{siunitx}
\usepackage{array}
\usepackage{multirow}
\usepackage{makecell}
\usepackage{amsmath}
\usepackage{cleveref}
\usepackage{verbatim}
\usepackage{amsfonts}
\usepackage{graphicx}
\usepackage{color}
\usepackage{float}
\usepackage{arydshln}

\def\ie{{\em i.e.,~}}
\def\eg{{\em e.g.,~}}

\usepackage[T1]{fontenc}

\usepackage[utf8]{inputenc}

\usepackage{microtype}

\usepackage{xcolor}
\newcommand{\dejiao}[1]{{\color{black}{#1}}}

\usepackage{hyperref}
\title{Learning Dialogue Representations from Consecutive Utterances}

\author{Zhihan Zhou\thanks{\; Work done during an internship at AWS AI Labs.} \quad Dejiao Zhang$^\dagger$ \quad Wei Xiao$^\dagger$ \quad Nicholas Dingwall$^\dagger$ \\ {\bf Xiaofei Ma}$^\dagger$ \quad {\bf Andrew O. Arnold}$^\dagger$ \quad {\bf Bing Xiang}$^\dagger$ \\
  $^\ast$Northwestern University \qquad $^\dagger$AWS AI Labs \\
 }

\begin{document}
\maketitle
\begin{abstract}
 
Learning high-quality dialogue representations is essential for solving a variety of dialogue-oriented tasks, especially considering that dialogue systems often suffer from data scarcity.
In this paper, we introduce Dialogue Sentence Embedding (DSE), a self-supervised contrastive learning method that learns effective dialogue representations suitable for a wide range of dialogue tasks. 
DSE learns from dialogues by taking consecutive utterances\footnote{Throughout this paper, we use \textit{utterance} to refer to all the sentences that belong to the same dialogue turn.} of the same dialogue as positive pairs for contrastive learning. Despite its simplicity, DSE achieves significantly better representation capability than other dialogue representation and universal sentence representation models.
We evaluate DSE on five downstream dialogue tasks that examine dialogue representation at different semantic granularities. 
Experiments in few-shot and zero-shot settings show that DSE outperforms baselines by a large margin. For example, it achieves $13\%$ average performance improvement over the strongest unsupervised baseline in 1-shot intent classification on 6 datasets.\footnote{The code and pre-trained models are publicly available at \url{https://github.com/amazon-research/dse}.}
We also provide analyses on the benefits and limitations of our model.

\end{abstract}

\section{Introduction}
Due to the variety of domains and the high cost of data annotation, labeled data for task-oriented dialogue systems is often scarce or even unavailable. 
Therefore, learning universal dialogue representations that effectively capture dialogue semantics at different granularities \citep{yutai, krone2020learning, yu2021few} provides a good foundation for solving various downstream tasks \citep{prototypical, vinyals2016matching}. 

Contrastive learning \citep{simclr, moco} has achieved widespread success in representations learning in both the image domain \cite{hjelm2018learning, imix, bachman2019learning} and the text domain \citep{simcse, zhang-etal-2021-pairwise,zhang-etal-2022-virtual,todbert}. Contrastive learning aims to reduce the distance between semantically similar (positive) pairs and increase the distance between semantically dissimilar (negative) pairs. These positive pairs can be either human-annotated or obtained through various data augmentations, while negative pairs are often collected through negative sampling in the mini-batch.

\begin{figure}[t]
		\centering
		\begin{tikzpicture}
		\draw (0,0 ) node[inner sep=0] {\includegraphics[width=1\columnwidth, trim={7cm 1.1cm 7cm 0.2cm}, clip]{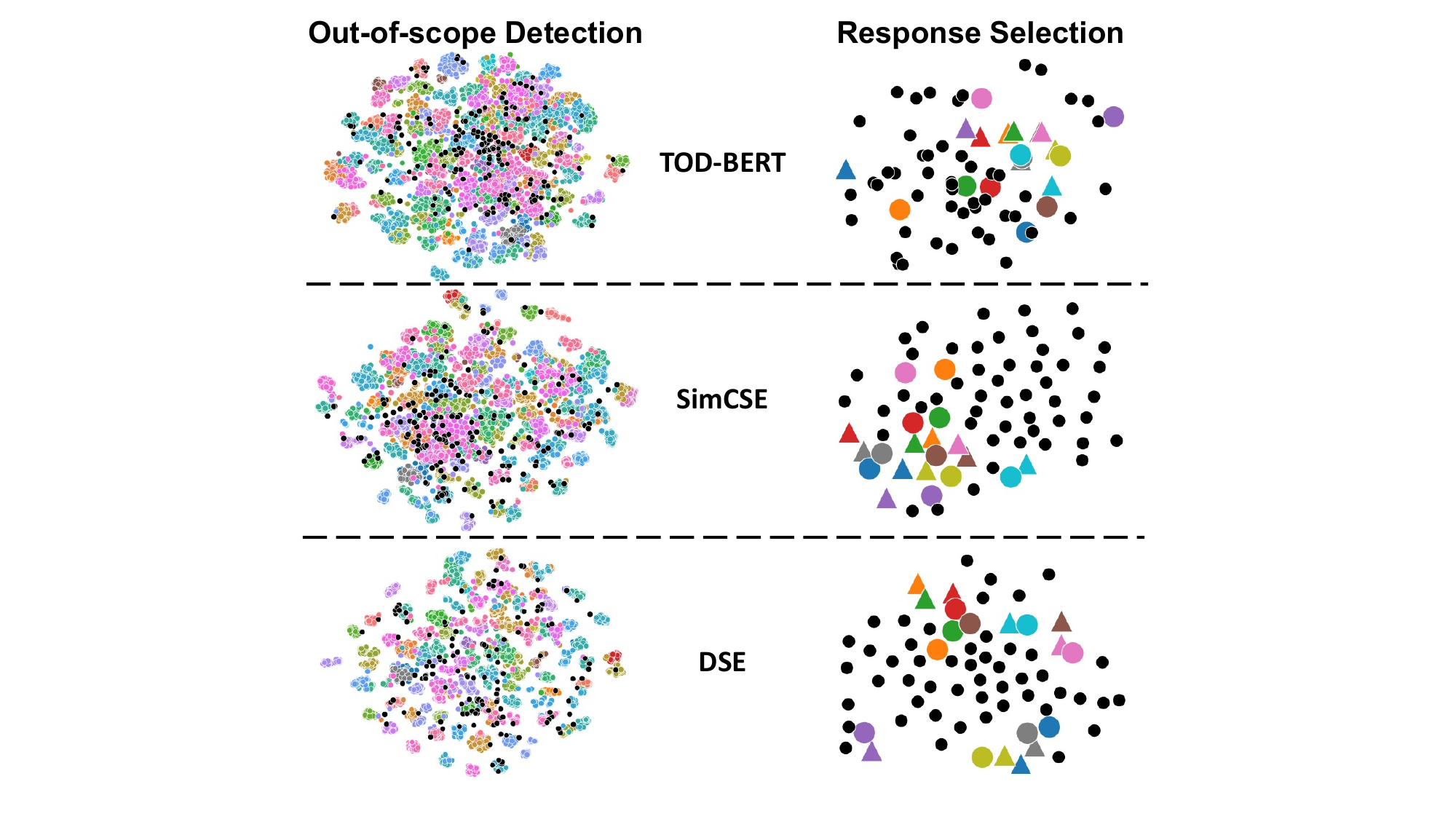}};
		\end{tikzpicture}
		\caption{TSNE visualization of the dialogue representations provides by \texttt{TOD-BERT},  \texttt{SimCSE}, and \texttt{DSE}. \textbf{Left}: each color indicates one intent category, while the black circles represents out-of-scope samples.
		\textbf{Right}: items with the same color stands for query-response pairs, where triangles represent queries. The black circles represents randomly sampled responses.
		}
		\label{fig:embedding}
\end{figure}

\begin{figure*}[t]
		\centering
		\begin{tikzpicture}
		\draw (0,0 ) node[inner sep=0] {\includegraphics[width=2\columnwidth, trim={1.1cm 9.7cm 0.7cm 0.8cm}, clip]{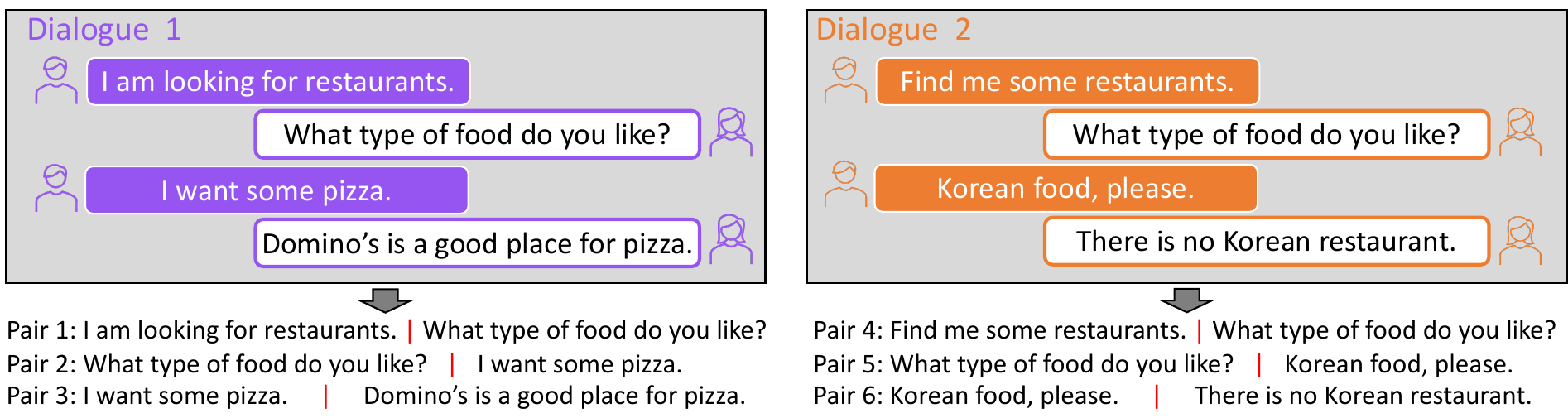}};
		\end{tikzpicture}
		\caption{Illustration of the positive pair construction from dialogues.}
		\label{fig:data_construction}
\end{figure*}

In the supervised learning regime,  \citet{simcse, zhang-etal-2021-pairwise} demonstrate the effectiveness of leveraging the Natural Language Inference (NLI) datasets \citep{nlidata-1, nlidata-2} to support contrastive learning. 
Inspired by their success, a natural choice of dialogue representation learning is utilizing the Dialogue-NLI dataset \citep{dianli} that consists of both semantically entailed and contradicted pairs.
However, due to its relatively limited scale and diversity, we found learning from this dataset leads to less satisfying performance, while the high cost of collecting additional human annotations precludes its scalability.
On the other extreme, unsupervised representation learning has achieved encouraging results recently, among which SimCSE \citep{simcse} and TOD-BERT \citep{todbert} set new state-of-the-art results on general texts and dialogues, respectively.

SimCSE uses Dropout \citep{dropout} to construct positive pairs from any text by passing a sentence through the encoder twice to generate two different embeddings. Although SimCSE outperforms common data augmentations that directly operate on discrete text, we find it performs poorly in the dialogue domain (see Sec. \ref{subsec:main_results}). This motivates us to seek better positive pair constructions by leveraging the intrinsic properties of dialogue data. On the other hand, TOD-BERT takes an utterance and the concatenation of all the previous utterances in the dialogue as a positive pair. 
Despite promising performance on same tasks, we found TOD-BERT struggles on many other dialogue tasks where the semantic granularities or data statistics are different from those evaluated in their paper.
 



In this paper, inspired by the fact that dialogues consist of consecutive utterances that are often semantically related, we use consecutive utterances within the same dialogue as positive pairs for contrastive learning (See Figure \ref{fig:data_construction}). 
This simple strategy works surprisingly well. 
We evaluate DSE on a wide range of task-oriented dialogue applications, including intent classification, out-of-scope detection, response selection, and dialogue action prediction. We demonstrate that DSE substantially outperforms TOD-BERT, SimCSE, and some other sentence representation learning models in most scenarios. 
We assess the effectiveness of our approach by comparing DSE against its variants trained on other types of positive pairs (e.g., Dropout and Dialogue-NLI).
We also discuss the trade-off in learning dialogue representation for tasks focusing on different semantic granularities and provide insights on the benefits and limitations of the proposed method.
Additionally, we empirically demonstrate that using consecutive utterances as positive pairs can effectively improve the training stability (Appendix \ref{subsec:epoch_results}).

\section{Why Contrastive Learning on Consecutive Utterances?}
When performing contrastive learning on consecutive utterances, we encourage the model to treat an utterance as similar to its adjacent utterances and dissimilar to utterances that are not consecutive to it or that belong to other dialogues. 


On the one hand, this training process directly increases an utterance's similarity with its \textit{true} response and decreases its similarities with other randomly sampled utterances. The ability to identify the appropriate response from many similar utterances is beneficial for dialogue ranking tasks (e.g., response selection). 
On the other hand, consecutive utterances also contain implicit categorical information, which benefits dialogue classification tasks (e.g., intent classification and out-of-scope detection).
Consider pairs 1 and 4 in Figure \ref{fig:data_construction}: we implicitly learn similar representations of \textit{I am looking for restaurants} and \textit{Find me some restaurants}, since they are both consecutive with \textit{What type of food do you like?}. 

In contrast, SimCSE does not enjoy these benefits by simply using Dropout as data augmentation. Although TOD-BERT also leverages the intrinsic dialogue semantics by combining an utterance with its dialogue context as positive pair, the context is often the concatenation of $5$ to $15$ utterances.
Due to the large discrepancy in both semantics and data statistics between each utterance and its context, simply optimizing the similarity between them leads to less satisfying representations on many dialogue tasks. As shown in Section \ref{sec:experiments}, TOD-BERT can even lead to degenerated representations on some downstream tasks when compared to the original BERT model. 




\section{Model}
\subsection{Notation}
Let $\{(x_i, x_{i^+})\}^M_{i=1}$ be a batch of positive pairs, where $M$ is the batch size. In our setting, each $(x_i, x_{i^+})$ denotes a pair of consecutive utterances sampled from a dialogue. Let $e_i$ denote the representation of the text instance $x_i$ that is obtained through an encoder. In this paper, we use mean pooling to obtain representations. 

\subsection{Training Target}
\label{sec:contrastive_loss}
Contrastive learning aims to maximize the similarity between positive samples and minimize the similarity between negative samples. For a contrastive anchor $x_i$, the contrastive loss aims to increase its similarity with its positive sample $x_{i^+}$ and decrease its similarity with the other $2M-2$ negative samples within the same batch. 

We adopt the \textit{Hard-Negative} sampling strategy proposed by \citet{zhang-etal-2021-pairwise}, which puts higher weights on the samples that are close to the anchor in the representation space. The underlying hypothesis is that hard negatives are more likely to occur among those that are located close to the anchor in the representation space. Specifically, the Hard-Negative sampling based contrastive loss regarding anchor $x_i$ is defined as follows:

\begin{equation}
\begin{split}
\ell^{i,i^+} =  
-\log \frac{\exp (\mathrm{sim}(e_i, e_{i^+}) / \tau)}
{\sum_{j \neq i} \exp(\alpha_{ij} \cdot \mathrm{sim}(e_i, e_j) / \tau)} \;. 
\end{split}
\label{contrastive_loss}
\end{equation}
As mentioned above, here $i$ and $i^+$ represent the indices of the anchor and its positive sample. We use $\tau$ to denote the temperature hyperparameter and $\mathrm{sim}(e_i, e_j)$ represent the cosine similarity of $e_i$ and $e_j$. 
In the above loss, $\alpha_{ij}$ is defined as follows, 
\begin{equation}
    \alpha_{ij} = \frac{\exp (\mathrm{sim}(e_i, e_{j}) / \tau)}{\frac{1}{2M-2} \sum_{k \neq {i^+}} \exp (\mathrm{sim}(e_i, e_{k}) / \tau)}\;. 
\end{equation}
Noted here, the denominator is averaged over all the other $2M$-2 negatives of $x_i$. Intuitively, samples that are close to the anchor in the representation space are assigned with higher weights. In other words, $\alpha_{ij}$ denotes the relative importance of instance $x_j$ for optimizing the contrastive loss of the anchor $x_i$ among all $2M$-2 negatives. For every positive pair $(x_i, x_{i^+})$, we respectively take $x_i$ and $x_{i^+}$ as the contrastive anchor to calculate the contrastive loss. Thereby, the contrastive loss over the batch is calculated as:

\begin{equation}
    \mathcal{L} = \frac{1}{2M} \sum_{i=1}^M ( \ell^{i,i^+} + \ell^{i^+,i}  )
\end{equation}

\dejiao{Here $\ell^{i^+,i}$ is defined by exchanging the roles
of instances $i$ and $i^+$ in Equation (\ref{contrastive_loss}), respectively.}

\begin{table*}[t]
	\centering
	\footnotesize
	\setlength{\tabcolsep}{2.4mm}{
	\begin{tabular}{lllr}\toprule
		
		 {\textbf{Task}} & {\textbf{Dataset}} & {\textbf{Evaluation Setting}} &  {\textbf{Num.}} \\
		 
		 \midrule
		 
		 \multirow{2}{*}{\textbf{Intent Classification}} & \texttt{Clinc150}, \texttt{Snips}, \texttt{Hwu64}, & 1-shot \& 5-shot fine-tune  & 10  \\ & \texttt{Bank77}, \texttt{Appen-A}, \texttt{Appen-H} & 1-shot \& 5-shot similarity & 10 \\
		 
		 \cmidrule(lr){1-4}

		 \textbf{Out-of-scope Detection} & \texttt{Clinc150} & 1-shot \& 5-shot similarity & 10 \\
		 
		 \cmidrule(lr){1-4}

		 \multirow{2}{*}{\textbf{Utterance-level Response Selection}} &  \multirow{2}{*}{\texttt{AmazonQA}} & 0-shot similarity& 1\\ && 500-shot \& 1000-shot fine-tune & 5  \\
		 
		 \cmidrule(lr){1-4}

 		 \textbf{Dialogue-level Response Selection} & \texttt{DSTC7-Ubuntu} &  0-shot similarity  & 1\\
 		 
 		 \cmidrule(lr){1-4}

   		 \textbf{Dialogue Action Prediction} & \texttt{DSTC2}, \texttt{GSIM} &  10-shot \& 20-shot fine-tune   &5 \\
		\bottomrule
	\end{tabular}}
	\caption{ \footnotesize  
		Summarization of all the experimental settings. Please see Appendix \ref{subsec:evaluation_data} for details of each dataset. The last column (\textbf{Num}) indicates the number of independent experiments with different random seeds (we report the averaged results). Since there is no randomness in 0-shot evaluations, we only run them once.
	}\label{tb:tasks}
\end{table*}

\section{Experiments}
\label{sec:experiments}
We run experiments with five different backbones:  BERT$_\texttt{base}$, BERT$_\texttt{large}$ \citep{bert}, RoBERTa$_\texttt{base}$, RoBERTa$_\texttt{large}$ \citep{roberta}, DistilBERT$_\texttt{base}$ \citep{distilbert}. Due to the space limit, we only present the results on BERT$_\texttt{base}$ in the main text. The results of other models are summarized in Appendix \ref{sec:other_backbone}. 
We use the same training data as TOD-BERT for a fair comparison.
We summarize the implementation details and data statistics of both pre-training and evaluation in Appendices \ref{sec:pre-train} and \ref{sec:evaluation_details}, respectively.

\subsection{Baselines}
We compare DSE against several representation learning models that attain state-of-the-art results on both general text and dialogue languages. We categorize them into the following two categories. 

\paragraph{Supervised Learning}{ 
\texttt{SimCSE-sup} \citep{simcse} is the supervised version of SimCSE, which uses entailment and contradiction pairs in the NLI datasets \citep{nlidata-1, nlidata-2} to construct positive pair and hard negative pair accordingly.  
In a similar vein, \texttt{PairSupCon} \cite{zhang-etal-2021-pairwise} leverages the entailment pairs as positive pairs only while proposing an unsupervised hard negative sampling strategy that we summarized in Section \ref{sec:contrastive_loss}. 
Following this line, we also evaluate DSE against its variant \texttt{DSE-dianli} trained on the Dialogue Natural Language Inference dataset \citep{dianli} by taking all the \textit{entail} pairs as positive pairs. }

\paragraph{Unsupervised Learning}
{
\texttt{TOD-BERT} \citep{todbert} optimizes a contrastive response selection objective by treating an utterance and its dialogue context as positive pair.
\texttt{DialoGPT} \citep{dialogpt} is a dialogue generation model that learns from consecutive utterance by optimizing a language modeling target.\footnote{We use mean pooling of its hidden states as sentence representation, which leads to better performance than using only the last token. We use its \textit{Medium} version that has twice as many parameters as \textbf{BERT$_{\texttt{base}}$}, since we found its \textit{Small} version performs dramatically worse under our settings.}
\texttt{SimCSE-unsup} \citep{simcse} uses Dropout \citep{dropout} to construct positive pairs. In the general text domain, \texttt{SimCSE-unsup} has attained impressive performance over several explicit data augmentation strategies that directly operate on the discrete texts. To test its effectiveness in the dialogue domain, we compare DSE against its variant \texttt{DSE-dropout} where augmentations of every single utterance are obtained through Dropout.

The evaluations on \texttt{DSE-dropout} and \texttt{DSE-dianli} allow us to fairly compare our approach against the state-of-the-art approaches in both the supervised learning and the unsupervised learning regimes.}

\subsection{Evaluation Setting} 
To accommodate the fact that obtaining a large number of annotations is often time-consuming and expensive for solving the task-oriented dialogue applications, especially considering the variety of domains and certain privacy concerns, we mainly focus on few-shot or zero-shot based evaluations.

\subsubsection{Evaluation Methods} 
Considering that only a few annotations are available in our setting, we mainly focus on the  \textbf{similarity-based} evaluations, where predictions are made based on different similarity metrics applied in the embedding space without requiring updating the model.

We use different random seeds to independently construct multiple (See Table \ref{tb:tasks}) few-shot train and validation sets from the original training data and use the original test data for performance evaluation. To examine whether the performance gap reported in the similarity-based evaluations is consistent with the associated fine-tuning approaches, we also report the \textbf{fine-tuning} results. We perform early stopping according to the validation set and report the testing performance averaged over different data splits.

\begin{table*}[ht]
	\centering
	\footnotesize
	\setlength{\tabcolsep}{2.4mm}{
	\begin{tabular}{llccccccc}\toprule
		 
		  & \textbf{BERT$_{\texttt{base}}$} & {\textbf{Clinc150}} & {\textbf{Bank77}} & {\textbf{Snips}}   & {\textbf{Hwu64}}  & {\textbf{Appen-A}} & {\textbf{Appen-H}} & {\textbf{Ave.}} \\

		\midrule
        
        \multirow{7}{*}{\rotatebox{90}{\large \texttt{1-shot}}} &
		{\textbf{SimCSE-sup}$^{\clubsuit}$ } & 52.30 & 38.05 &	65.98 &	40.79 & 35.35	& 44.81 & 46.21  \\
		
		& {\textbf{PairSupCon}$^{\clubsuit}$ } & 55.34 & 41.30 &	65.20 & 41.43 & 37.55	& 47.55 & 48.06 \\
		
        & {\textbf{DSE-dianli$^{\clubsuit}$ (ours)}}  & 45.91&	38.33&	58.23&	34.95&	33.87&	42.26&	42.26\\
		
		\cmidrule(lr){2-9}

		& {\textbf{BERT}$^{\diamondsuit}$ } &36.98&	22.05 &	62.51 &	27.74 &	13.19 &	18.74 & 30.20\\

		& {\textbf{SimCSE-unsup}$^{\diamondsuit}$}  &46.44&	37.51&	59.58&	34.34&	27.10	&36.00 & 40.16\\
		
		& {\textbf{DialoGPT}$^{\diamondsuit}$}  & 42.23 & 28.08 & 63.10 & 30.45 & 18.90 & 24.48 & 34.54 \\
		
		& {\textbf{TOD-BERT}$^{\diamondsuit}$}  &36.67&	27.11&	62.52&	29.52&	20.61& 	26.68 & 33.85 \\
		
		& {\textbf{DSE-dropout$^{\diamondsuit}$ (ours)}}  &46.48	 &30.02	 &65.03	 &33.25	 &16.94	 &21.77	 &35.58\\

		& {\textbf{DSE$^{\diamondsuit}$ (ours)} }& \textbf{62.53} &	\textbf{43.12} &	\textbf{79.57} &	\textbf{44.31} &\textbf{37.97} &	\textbf{48.71} &	\textbf{52.70} \\
		
	    \hdashline
	    \\
    \multirow{7}{*}{\rotatebox{90}{\large \texttt{5-shot}}} &
     {\textbf{SimCSE-sup}$^{\clubsuit}$ }   & 71.11&	56.38&	79.98&	56.52&	49.71	&59.42 &62.18 \\
		
		& {\textbf{PairSupCon}$^{\clubsuit}$ }    & 73.88&	60.07&	76.14&	55.75&	\textbf{52.71}	&62.23&63.46 \\
		
		& {\textbf{DSE-dianli$^{\clubsuit}$ (ours)}} & 60.65 &	49.78&	73.80&	46.65&	46.52&	54.39&	55.30\\
		  
		\cmidrule(lr){2-9}

		& {\textbf{BERT}$^{\diamondsuit}$ } & 59.48&	38.73&	78.65&	43.15&	21.39&	27.61 &44.83\\

		& {\textbf{SimCSE-unsup}$^{\diamondsuit}$}  & 65.37&	55.03&	77.01&	48.79&	43.35&	51.55 & 56.85\\
		
		& {\textbf{DialoGPT}$^{\diamondsuit}$}  &64.53 & 46.56 & 82.15 & 45.67 & 33.67 & 39.61 & 52.03 \\
		
		& {\textbf{TOD-BERT}$^{\diamondsuit}$}  &57.74	&42.98&	79.68&	42.32&	33.58&	42.52 &49.80\\
		
		& {\textbf{DSE-dropout$^{\diamondsuit}$ (ours)}} & 70.46&	49.95&	80.10&	52.16&	30.00&	37.48&	53.36\\

		& {\textbf{DSE$^{\diamondsuit}$ (ours)} }&  \textbf{78.73}&	\textbf{61.65}&	\textbf{88.62}&	\textbf{60.87}&	52.32&	\textbf{62.68}&	\textbf{67.48}\\

		\bottomrule
	\end{tabular}}
	\caption{ \footnotesize  
		Results on similarity-based 1-shot and 5-shot Intent Classification. Predictions are made purely based on the embeddings provided by each model without any parameter tuning. All the models use BERT$_{\texttt{base}}$ as the backbone model. 
		$\clubsuit$: Supervised models. $\diamondsuit$: Unsupervised models.
	}\label{tb:bert_intent_sim}
\end{table*}

\begin{table*}[ht]
	\centering
	\footnotesize
	\setlength{\tabcolsep}{2.4mm}{
	\begin{tabular}{llccccc}\toprule
		
		 &\textbf{BERT$_{\texttt{base}}$} &  {\textbf{Accuracy}} & {\textbf{In-Accuracy}} & {\textbf{OOS-Accuracy}}   & {\textbf{OOS-Recall}} & {\textbf{Ave.}}  \\

		\midrule
		\multirow{7}{*}{\rotatebox{90}{\large \texttt{m-d}}} &
		{\textbf{SimCSE-sup}$^{\clubsuit}$  }  & 44.63 & 	51.50 & 	78.63 & 	13.70 &47.12\\
		
		&{\textbf{PairSupCon}$^{\clubsuit}$ }   &51.87&	54.34&	82.33&	40.75	&57.32 \\
		
		&{\textbf{DSE-dianli$^{\clubsuit}$ (ours)}}  &44.73 &	44.88 &	80.83 &	44.07 &	53.63 \\
		
		\cmidrule(lr){2-7}
		
		&{\textbf{BERT}$^{\diamondsuit}$ } & 33.96&	36.01&	80.58&	24.77 &43.83 \\

		&{\textbf{SimCSE-unsup}$^{\diamondsuit}$}  &40.45	&45.50&	77.83&	17.73 &45.38\\
		
		& {\textbf{DialoGPT}$^{\diamondsuit}$}  & 36.98 & 40.70 & 80.73 & 20.21 & 44.66 \\
		
		&{\textbf{TOD-BERT}$^{\diamondsuit}$}  & 34.77&	36.28&	79.74&	27.98 &44.69\\
    
        &{\textbf{DSE-dropout}$^{\diamondsuit}$}  & 42.41 &	45.19 &	81.26 &	29.92 &	49.70\\

		&{\textbf{DSE$^{\diamondsuit}$ (ours)} } & \textbf{58.74}&	\textbf{60.52}&	\textbf{84.07}&	\textbf{50.72}&	\textbf{63.51} \\

	    \hdashline
	    \\

		\multirow{7}{*}{\rotatebox{90}{\large \texttt{mean}}} &
		{\textbf{SimCSE-sup}$^{\clubsuit}$  }  & 36.90	&29.07	&47.04	&72.12 &46.28\\
		
		&{\textbf{PairSupCon}$^{\clubsuit}$ }   &47.29	&37.44&	58.72&	91.63&58.77	 \\
		
		&{\textbf{DSE-dianli$^{\clubsuit}$ (ours)}}  &40.70 &	30.78 &	57.67 &	85.30 &	53.61 \\
		
		\cmidrule(lr){2-7}
		
		&{\textbf{BERT}$^{\diamondsuit}$ } & 35.64&	24.78&	53.09&	84.47 &49.50 \\

		&{\textbf{SimCSE-unsup}$^{\diamondsuit}$}  &37.65	&28.99	&49.38&	76.62&48.16 \\
		
		& {\textbf{DialoGPT}$^{\diamondsuit}$}  & 38.04 & 27.00 & 52.87 & 87.75 & 51.42 \\
		
		&{\textbf{TOD-BERT}$^{\diamondsuit}$}  & 36.31&	25.76&	53.40&	83.75 &49.81\\

		&{\textbf{DSE-dropout$^{\diamondsuit}$ (ours)}}  &41.19&	30.93&	54.76&	87.39	&53.57 \\

		&{\textbf{DSE$^{\diamondsuit}$ (ours)} } & \textbf{50.88}	&\textbf{41.72}	&\textbf{60.64}&	\textbf{92.11}&	\textbf{61.34} \\
		\bottomrule
	\end{tabular}}
	\caption{ \footnotesize  
		Results on similarity-based 1-shot out-of-scope detection on \texttt{Clinc150} dataset. 
		The out-of-scope threshold is respectively set as \textit{mean} (\texttt{m}) and \textit{mean-std} (\texttt{m-d}) of each sample's similarity with its closest category. See Sec. \ref{subsec:task_metric} for details. $\clubsuit$: Supervised models. $\diamondsuit$: Unsupervised models.
	}\label{tb:bert_oos_sim_1}
\end{table*}

\subsubsection{Tasks and Metrics}
\label{subsec:task_metric}
We evaluate all models considered in this paper on two types of tasks: \textbf{utterance-level} and \textbf{dialogue-level}. 
The utterance-level tasks take a single dialogue utterance as input, while the dialogue-level tasks take the dialogue history as input. 
These two types of tasks assess representation quality on dialogue understanding at different semantic granularities, which are shared across a variety of downstream tasks.

\paragraph{Intent Classification} is an utterance-level task that aims to classify user utterances into one of the pre-defined intent categories. We use Prototypical Networks \citep{prototypical} to perform the similarity-based evaluation. Specifically, we calculate a prototype embedding for each category by averaging the embedding of all the training samples that belong to this category. A sample is classified into the category whose prototype embedding is the most similar to its own. We report the classification accuracy for this task.

\paragraph{Out-of-scope Detection} advances intent classification by detecting whether the sample is out-of-scope, \ie does not belong to any pre-defined categories. We adapt the aforementioned Prototypical Networks to solve it. For a test sample, if its similarity with its most similar category is lower than a threshold, we classify it as out-of-scope. Otherwise, we assign it to its most similar category. For each model, we calculate the \texttt{mean} and \texttt{std} (standard deviation) of the similarity scores between every sample and its most similar prototype embedding, and take \texttt{mean}$-$\texttt{std} and \texttt{mean} as the threshold, respectively. The evaluation set contains both in-scope and out-of-scope examples. We evaluate this task with four metrics: 1) \underline{Accuracy}: accuracy of both in-scope and out-of-scope detection. 2)  \underline{In-Accuracy}: accuracy reported on 150 in-scope intents. 3) \underline{OOS-Accuracy}: out-of-scope detection accuracy. 4) \underline{OOS-Recall}: recall of OOS detection. 

\paragraph{Utterance-level Response Selection} is an utterance-level task that aims to find the most appropriate response from a pool of candidates for the input user query, where both the query and response are single dialogue utterances. We formulate it as a ranking problem and evaluate it with Top-k-100 accuracy (a.k.a., k-to-100 accuracy), a standard metric for this ranking problem \citep{todbert}. 
For every query, we combine its ground-truth response with 99 randomly sampled responses and rank these 100 responses based on their similarities with the query in the embedding space.
The Top-k-100 accuracy represents the ratio of the ground-truth response being ranked at top-k, where k is an integer between 1 and 100. We report the Top-1, Top-3, and Top-10 accuracy of the models.

\paragraph{Dialogue-Level Response Selection} is a dialogue-level task. The only difference with the \textit{Utterance-level Response Selection} is that query in this task is dialogue history (e.g., concatenation of multiple dialogue utterances from different speakers). We also report the Top-1, Top-3, and Top-10 accuracy for this task.

\paragraph{Dialogue Action Prediction} is a dialogue-level task that aims to predict the appropriate system action given the most recent dialogue history. We formulate it as a multi-label text classification problem and evaluate it with model fine-tuning. We report the Macro and Micro F1 scores for this task.


\subsection{Main Results}
\label{subsec:main_results}

\begin{table*}[ht]
	\centering
	\footnotesize
	\setlength{\tabcolsep}{2.4mm}{
	\begin{tabular}{lcccccc}\toprule
		
		\multirow{2}{*}{\textbf{BERT$_{\texttt{base}}$}} &
		\multicolumn{3}{c}{\textbf{AmazonQA}} & 
		\multicolumn{3}{c}{\textbf{DSTC7-Ubuntu}}  \\
		\cmidrule(lr){2-4} \cmidrule(lr){5-7}
		& {\textbf{Top-1 Acc.}} & {\textbf{Top-3 Acc.}} & {\textbf{Top-10 Acc.}} & {\textbf{Top-1 Acc.}} & {\textbf{Top-3 Acc.}} & {\textbf{Top-10 Acc.}} \\

		\midrule
		{\textbf{SimCSE-sup}$^{\clubsuit}$  }  &47.03&	62.40&76.80 &11.37&	19.40	&33.53 \\
		
		{\textbf{PairSupCon}$^{\clubsuit}$ }   & 52.22&	65.09&	76.85& \textbf{15.00}&	23.02&	\textbf{35.73}\\
		
        {\textbf{DSE-dianli$^{\clubsuit}$ (ours)}}  & 49.16&	63.36&	76.66&	14.92&	22.73	&34.72\\

		\midrule
		
		{\textbf{BERT}$^{\diamondsuit}$ } & 29.70&	43.86&	60.36 &6.75	&12.97&	24.20 \\

		{\textbf{SimCSE-unsup}$^{\diamondsuit}$}  & 48.02&	62.45&	76.00 &10.03	&17.13&	29.37\\
		
		{\textbf{DialoGPT}$^{\diamondsuit}$}  & 35.96 & 49.52 & 64.44 & 10.20 & 17.60 & 29.82 \\
		
		{\textbf{TOD-BERT}$^{\diamondsuit}$} & 27.25 &40.26&	56.63&5.52&	10.55&	22.30  \\
        
        {\textbf{DSE-dropout$^{\diamondsuit}$ (ours)}}  & 37.80	&51.64&	66.58&9.55&	16.97&	28.80\\
        
		{\textbf{DSE$^{\diamondsuit}$ (ours)} } &	\textbf{56.62}&		\textbf{70.54}&		\textbf{81.90}&14.78&	\textbf{23.10}&	\textbf{35.73} \\

		\bottomrule
	\end{tabular}}
	\caption{ \footnotesize  
		Results on 0-shot response selection on \texttt{AmazonQA} (utterance-level) and \texttt{DSTC7-Ubuntu} (dialogue-level).
	}\label{tb:bert_rs_ubuntu_sim}
\end{table*}

\begin{table*}[ht]
	\centering
	\footnotesize
	\setlength{\tabcolsep}{2.4mm}{
	\begin{tabular}{lccccc}\toprule
		\multirow{2}{*}{\textbf{BERT$_{\texttt{base}}$}} &
		\multicolumn{2}{c}{\textbf{DSTC2}} & 
		\multicolumn{2}{c}{\textbf{GSIM}}  \\
		\cmidrule(lr){2-3} \cmidrule(lr){4-5}
		& {\textbf{10-shot}} & {\textbf{20-shot}} & {\textbf{10-shot}}   & {\textbf{20-shot}} & \textbf{Ave.} \\

		\midrule
		{\textbf{SimCSE-sup}$^{\clubsuit}$  }  & 84.12 || 36.62& 	86.15 || 36.99& 	77.22 || 35.03& 	84.75 || 38.67 & 59.94 \\
		
		{\textbf{PairSupCon}$^{\clubsuit}$ }   & 84.42 || 36.52& 	86.22 || 36.87& 	74.35 || 33.44& 	82.26 || 37.62	&58.96 \\
		
		{\textbf{DSE-dianli$^{\clubsuit}$ (ours)}}  &83.99 || 	36.20&	86.74 || 	37.02	&69.52 || 	31.36&	79.97 || 	36.62 & 57.68\\
		
		\midrule
		
		{\textbf{BERT}$^{\diamondsuit}$ } & 81.74 || 34.78& 	86.98 || 37.28& 	70.67 || 31.24& 	77.60 || 35.74  & 57.00\\

		{\textbf{SimCSE-unsup}$^{\diamondsuit}$}  & 84.41 || 36.62& 	87.84 || 37.98& 	75.78 || 34.47& 	81.73 || 37.64 & 59.56\\
		
		{\textbf{TOD-BERT}$^{\diamondsuit}$}  & 87.12 || 36.83& 	88.59 || 37.90& 	85.63 || 38.53& 	92.15 || 42.04 & 63.60 \\
		
		{\textbf{DSE-dropout$^{\diamondsuit}$ (ours)}}  &83.23 || 36.18&	86.65 || 36.95&	72.25 || 32.70	&81.91 || 37.33 & 58.62\\

		{\textbf{DSE$^{\diamondsuit}$ (ours)} } & 84.58 || 36.02 & 88.01 || 38.01 & 79.26 || 35.89 & 86.73 || 39.51 & 61.03 \\
		
		\hdashline
		
		{\textbf{DSE$_{2\text{-}1}$ (ours)}} & 84.47 || 	36.09&	88.86 || 	38.41&	83.81	 || 37.78&	88.03 || 	40.29 & 62.22\\
		
		{\textbf{DSE$_{3\text{-}1}$ (ours)}}  & 88.78 || 	38.52&	89.59 || 	38.58&	85.27	 || 39.10&	88.65 || 	40.87 & 63.67\\
		
		{\textbf{DSE$_{123\text{-}1}$ (ours)}} & \textbf{89.48} || 	\textbf{38.60}&	\textbf{90.97} || 	\textbf{39.79}&	\textbf{87.90} || 	\textbf{40.05}&	\textbf{92.48} || 	\textbf{42.22} & \textbf{65.19} \\

		\bottomrule
	\end{tabular}}
	\caption{ \footnotesize  
		Results on 10-shot and 20-shot dialogue action prediction fine-tuning on \texttt{DSTC2} amd \texttt{GSIM}. We use "||" to separate the Micro F1 score and Macro F1 score. $\clubsuit$: Supervised models. $\diamondsuit$: Unsupervised models.
	}\label{tb:bert_da_ft}
\end{table*}

\paragraph{Intent Classification \& Out-of-scope Detection} Tables \ref{tb:bert_intent_sim} and \ref{tb:bert_oos_sim_1} show the results of similarity-based intent classification and out-of-scope detection. The fine-tuning based results are presented in Appendix \ref{sec:bertbase_other}. As we can see, DSE substantially outperforms all the baselines. In intent classification, it attains $13\%$ average accuracy improvement over the strongest unsupervised baseline. More importantly, DSE achieves a $5\%$--$10\%$ average accuracy improvement over the supervised baselines that are trained on a large amount of expensively annotated data. The same trend was observed in out-of-scope detection, where DSE achieves $13\%$-$20\%$ average performance improvement over the strongest unsupervised baseline. The comparison between DSE, DSE-dropout, and DSE-dianli further demonstrates the effectiveness of using consecutive utterances as positive pairs in learning dialogue embeddings.

The left panel of Figure \ref{fig:embedding} visualizes the embeddings on the \texttt{Clinc150} dataset given by  TOD-BERT, SimCSE, and DSE, which provides more intuitive insights into the performance gap. 
As shown in the figure, with the DSE embeddings, in-scope samples belonging to the same category are closely clustered together. Clusters of different categories are clearly separated with a large margin, while the out-of-scope samples are far away from those in-scope clusters.

\paragraph{Response Selection} 
Table \ref{tb:bert_rs_ubuntu_sim} shows the the results of similarity-based 0-shot response selection on utterance-level (AmazonQA) and dialogue-level (DSTC7-Ubuntu). Results of finetune-based evaluation on AmazonQA show similar trend and we summarize in Table \ref{tb:bert_rs_amazon_ft} in Appendix. In Table \ref{tb:bert_rs_ubuntu_sim}, 
the large improvement attained by DSE over the baselines indicate our model's capability in dialogue response selection, in presence of both single-utterance query or using long dialogue history as query. 
The right panel of Figure \ref{fig:embedding} further illustrates this. It visualizes the embeddings of questions and answers in the AmazonQA dataset calculated by DSE, SimCSE, and TOD-BERT. With the DSE embedding, 
each question is placed close to its real answer while far away from other candidates.

\paragraph{Dialogue Action Prediction} 
Table \ref{tb:bert_da_ft} shows that DSE outperforms all baselines except TOD-BERT, which indicates its capability in capturing dialogue-level semantics. To better understand TOD-BERT's superiority over DSE on this task, we further investigate this task and find its data format is special. Concretely, here each input consists of multiple utterances explicitly concatenated by using two special tokens \textsc{[SYS]} and \textsc{[USR]} to indicate the system and user inputs, respectively. For example, $(\textsc{[SYS]} \; \text{hi} \; \textsc{[USR]} \; \text{how are you?} \; \textsc{[SYS]} \; \text{I'm good})$. It follows the same format as the queries\footnote{We use \textit{query} to refer the first utterance in a positive pair and use \textit{response} to refer the other one} used for training  TOD-BERT, while DSE uses a single utterance as the query.

\begin{table}[ht]
	\centering
	\footnotesize
	\begin{tabular}{lrrrrr}
		\toprule
		 \textbf{BERT$_{\texttt{base}}$} & \textbf{IC} & \textbf{OOS} & \textbf{u-RS} & \textbf{d-RS} & \textbf{DA} \\
		\midrule
		
		\textbf{TOD-BERT} & 41.83 &47.25 & 41.38 & 12.79 & 63.60 \\
		\textbf{DSE}  & \textbf{60.09} & \textbf{62.43} & \textbf{69.69} &  \textbf{24.54} & 61.03 \\

		\textbf{DSE$_{2\text{-}1}$} & 56.56 & 61.55 & 59.88 & 19.36  & 62.22 \\ 
		\textbf{DSE$_{3\text{-}1}$}  & 57.26 & 61.19 & 61.94 & 22.04  & 63.67  \\ 
		\textbf{DSE$_{123\text{-}1}$} & 59.60  & 61.59 & 63.67 &  22.63 & \textbf{65.19} \\

		\bottomrule
	\end{tabular}
	\caption{
		\footnotesize
		Performance of TOD-BERT, DSE, and its variants on intent classification (IC), out-of-scope detection(OOS), response selection on utterance-level (u-RS) and dialogue-level (d-RS), dialogue action prediction (DA).
	}\label{tb:bertbase_ablation}
\end{table}

\subsection{Trade-off in Query Construction}
To understand the impact of using multiple utterances as queries, we train three new variants of DSE. 
Specifically, we construct positive pairs as: $(u_1 \; \textsc{[SEP]} \; u_2, u_3)$, $(u_2 \; \textsc{[SEP]} \; u_3, u_4)$, where $u_i$ represents the $i$-th utterance in a dialogue. We use the \textsc{[SEP]} token to concatenate two consecutive utterances as query. We refer DSE trained with this data as \texttt{DSE$_{2\text{-}1}$} since it uses $2$ utterances as the query and $1$ utterance as the response. Similarly, we train another variant \texttt{DSE$_{3\text{-}1}$}. Lastly, we also combine the positive pairs constructed for training DSE, DSE$_{2\text{-}1}$, and DSE$_{3\text{-}1}$ together to train another variant named \texttt{DSE$_{123\text{-}1}$}.

As shown in Table \ref{tb:bert_da_ft}, by simply increasing the number of utterances within each query to three, DSE again outperforms TOD-BERT, and the improvement further expands when trained with the combined set, \ie \texttt{DSE$_{123\text{-}1}$}. Our results demonstrate that using long queries that consist of $5$ to $15$ utterances as what TOD-BERT does is not necessary even for dialogue action prediction. 
We further demonstrate this by evaluating DSE and its variants on all the other four tasks in Table \ref{tb:bertbase_ablation}, where our model outperforms TOD-BERT by a large margin. 
As it indicates, by using a single utterance as a query, DSE achieves a good balance among different dialogue tasks. 
In cases where dialogue action prediction is of great importance, augmenting the original training set of DSE with positive pairs constructed by using query consisting of 2 to 3 utterances is good enough to attain better performance while only incurring a slight performance drop on other tasks.

\subsection{Potential Limitation}
\label{subsec:limitation}
Considering the effectiveness of using consecutive utterances as positive pairs, a natural yet important question is: what are the potential limitations of our proposed approach? When using consecutive utterances as positive pairs for contrastive learning, an assumption is that responses to the same query are semantically similar. Vice versa, queries that prompt the same answer are similar. This assumption holds in many scenarios, yet it fails sometimes. 

It may fail when answers have different semantic meanings. Take the pairs 2 and 5 in Figure \ref{fig:data_construction} as an example. Through our data construction, we implicitly consider \textit{I want some pizza} and \textit{Korean food, please} to be semantically similar since they are both positively paired with \textit{What type of food do you like}. 
Although this may be correct in some coarse-grained classification tasks since these two sentences generally represent the same intent (e.g., order food), 
using them as positive pairs can introduce some noise when considering more fine-grained semantics.
This problem is further elaborated when answers are general and ubiquitous, \eg \textit{Thank you}. Since these utterances can be used to respond to countless types of dissimilar queries, \eg \textit{I have booked a ticket for you} v.s. \textit{Happy birthday}, we may implicitly increase the similarities among highly dissimilar utterances when training on these samples, which is undesirable. 

We verify this on the NLI datasets,  where the
the task is to identify whether one sentence semantically entails or contradicts the anchor sentence. For each anchor sentence, we calculate its cosine similarities with both the true \textit{entailment}, \textit{contradiction} sentences in the representation space. We classify the sentence with higher cosine similarity with the anchor as entailment and the other as the contradiction. 
Despite DSE achieves better classification accuracy (76.62) than BERT (69.40) and TOD-BERT (70.51), it underperforms SimCSE-unsup (80.31). Although using dropout to construct positive pairs is not as effective as ours in many dialogue scenarios, this method better avoids introducing fine-grained semantic noise.



Despite the limitations, using consecutive utterances as positive pairs still leads to better dialogue representation than the elaborately labeled NLI datasets, indicating the great value of the information contained in dialogue utterances.


		
		
		

		
        


\section{Related Work}

\paragraph{Positive Pair Construction} Popular supervised sentence representation learning often takes advantage of the human-annotated natural language inference (NLI) datasets \citep{nlidata-1, nlidata-2} for contrastive learning \citep{simcse, zhang-etal-2021-pairwise, sentencebert, sentenceencoder}. These sentence pairs either entail or contradict each other, making them the great choice for constructing positive and negative training pairs.
Unsupervised sentence representation learning often relies on variant data augmentation strategies.
\citet{logeswaran2018efficient} and \citet{declutr} propose using sentences and their surrounding context as positive pairs. Other works resort to popular NLP augmentation methods such as word permutation \citep{wu2020clear} and back-translation \citep{fang2020cert}. Recently, \citet{simcse} demonstrates the superiority of using Dropout over other data augmentations that directly operate on the discrete texts.

\paragraph{Contrastive Learning Methods} Contrastive learning is key to recent advances in learning sentence embeddings. Many contrastive learning approaches utilize memory-based methods, which draw negative samples from a memory bank of embeddings \citep{hjelm2018learning, bachman2019learning, moco}. On the other hand, \citet{simclr} introduces a memory-free contrastive framework, SimCLR, that takes advantage of negative sampling within large mini-batches. Promising results were also reported in the NLP domain. To name a few, \citet{simcse} leverages both within batch negatives and the `contradiction' annotations in NLI; and \citet{zhang-etal-2021-pairwise} propose an unsupervised hard-negative sampling strategy.

\paragraph{Dialogue Language Model} 
Learning dialogue-specific language models has attracted a lot of attention.
Along this line, \citet{dialogpt} adapts the pre-trained GPT-2 model \citep{gpt2} on Reddit data to perform open-domain dialogue response generation. \citet{plato} evaluates multiple dialogue generation tasks after training on Twitter and Reddit data \citep{wolf2019transfertransfo, peng2020few}. 
For dialogue understanding, \citet{henderson2019training} propose a response selection approach using a dual-encoder model. They pre-train the response selection model on Reddit and then fine-tune it for different response selection tasks. Following this, \citet{convert} introduces a more efficient conversational model that is pre-trained with a response selection target on the Reddit corpus. 
However, they did not release code or pre-trained models for comparison.
\citet{todbert} combines nine dialogue datasets to obtain a large and high-quality task-oriented dialogue corpus. They introduce the TOD-BERT model by further pre-training BERT on this corpus with both the masked language modeling loss and the contrastive response selection loss.

\section{Conclusion}
In this paper, we introduce a simple contrastive learning method DSE that learns dialogue representations by leveraging consecutive utterances in dialogues as positive pairs. 
We conduct extensive experiments on five dialogue tasks to show that the proposed method greatly outperforms other state-of-the-art dialogue representation models and universal sentence representation methods. 
We provide ablation study and analysis on our proposed data construction from different perspectives, investigate the trade-off between different data construction variants, and discuss the potential limitation to motivate further exploration in representation learning on unlabeled dialogues.
We believe DSE can serve as a drop-in replacement of the dialogue representation model (e.g., the text encoder) for a wide range of dialogue systems. 


\bibliography{custom}
\bibliographystyle{acl_natbib}

\newpage
\appendix

\section{Pre-train} 
\label{sec:pre-train}
In this section, we present the training data, implementation details, and training stability of model pre-training. 

\subsection{Data} We utilize the corpus collected by TOD-BERT \citep{todbert} to construct positive pairs. This dataset is the combination of $9$ publicly available task-oriented datasets: MetaLWOZ \citep{MetaLWOZ}, Schema \citep{Schema}, Taskmaster \citep{Taskmaster}, MWOZ \citep{MWOZ}, MSR-E2E \citep{MSR-E2E}, SMD \citep{SMD}, Frames \citep{Frames}, WOZ \citep{WOZ}, CamRest676 \citep{CamRest676}. The combined dataset contains 100707 dialogues with 1388152 utterances over 60 domains. We filter out sentences with less or equal to 3 words and end up with 892835 consecutive utterances (for DSE) and 879185 unique sentences (for DSE-dropout). Note that, the training data of SimCSE-unsup consists of 1 million sentences from Wikipedia. That says, on the one hand, we use the same dataset as TOD-BERT but with our proposed data construction. On the other hand, we use a similar number of training samples as SimCSE-unsup. We believe such data construction makes the comparisons fair enough.

\subsection{Hyperparameters} We add a contrastive head after the Transformer model and use the outputs of the contrastive head to perform contrastive learning. We use a two-layer MLP with size ($d \times d, d \times 128$) as the contrastive head. We use Adam \citep{adam} with a batch size of $1024$ and a constant learning rate as the optimizer. We set the learning rate for contrastive head as $3e-4$ and the learning rate for the Transformer model as $3e-6$. The temperature hyperparameter $\tau$ is set as 0.05. We train the model for $15$ epochs (see Appendix \ref{subsec:epoch_results} for more details) and save the model at the end for evaluation. We use the same hyperparameters across all the experiments for BERT$_\texttt{base}$, RoBERTa$_\texttt{base}$, and DistilBERT$_\texttt{base}$ models. For BERT$_\texttt{large}$ and RoBERTa$_\texttt{large}$, we change the batch size to $512$ to fit it into the GPUs. Pre-training of the DistilBERT$_\texttt{base}$, BERT$_\texttt{base}$, and BERT$_\texttt{large}$ model respectively takes $3$, $4$, and $13$ hours on $8$ NVIDIA® V$100$ GPUs\footnote{Our codes and model are under the Apache-2.0 License}.

\subsection{Training Stability}
\label{subsec:epoch_results}
\begin{figure}[t]
		\centering
		\begin{tikzpicture}
		\draw (0,0 ) node[inner sep=0] {\includegraphics[width=1\columnwidth, trim={1.8cm 0cm 1.5cm 1cm}, clip]{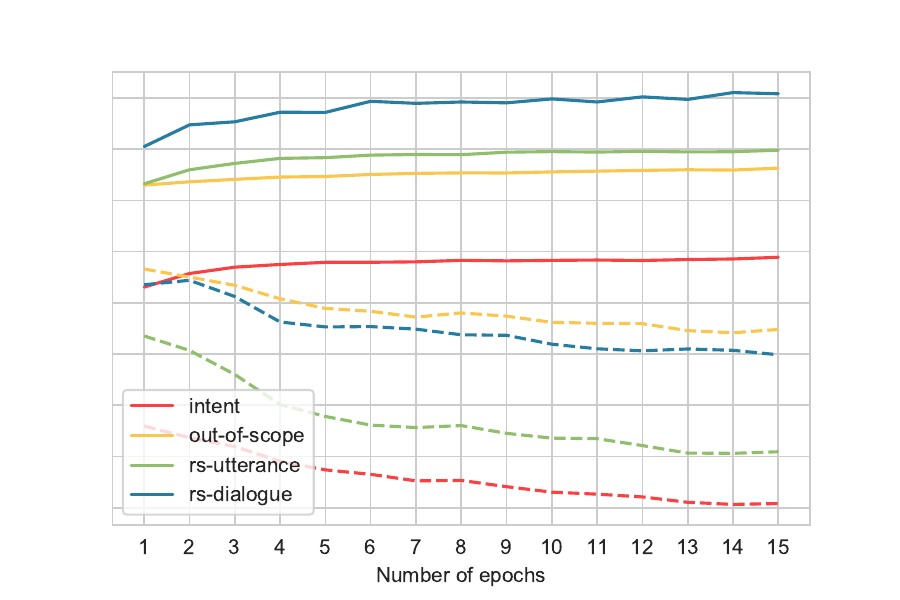}};
		\end{tikzpicture}
		\caption{\texttt{DSE} and \texttt{DSE-dropout}'s performance on each task at every epoch. The dashed lines represent the results of \texttt{DSE-dropout}.}
		\label{fig:epoch}
\end{figure}

In this section, we analyze the model's stability in terms of training steps when training with different type of positive pairs. We compare two data construction methods: consecutive utterances (DSE) and dropout (DSE-dropout). We first train each model for $15$ epochs, save the checkpoint at the end of each epoch and evaluate each checkpoint with similarity-based methods. Figure \ref{fig:epoch} shows the two model's average performances on intent classification, out-of-scope detection, utterance-level response selection and dialogue-level response selection.

This result further illustrates the effectiveness of using consecutive utterances as positive pairs for learning dialogue representation.
As shown in the figure, DSE's performance on all the tasks consistently improves during the training process, while DSE-dropout achieves the best performance at the first epoch and significantly loses performance afterwards. Besides, DSE's performance is less sensitive to the training steps. It achieves stable performance after about 5 epochs. In contrast, DSE-dropout's performance drops dramatically during the training process, yet it never surpasses DSE's performance. Therefore, we report DSE-dropout's performance at the first epoch in all the tables.

\section{Evaluation Setup}
\label{sec:evaluation_details}
In this section, we present evaluation details and introduction to the evaluation dataset. Throughout this paper, we use \textit{cosine similarity} as the similarity metric and \textit{mean pooling of token embeddings} as the sentence representation. For baseline models, we report the better results of using its default setting (e.g., last hidden state of the [CLS] token as sentence embedding for SimCSE) and mean pooling.

\subsection{Hyperparameters}
We use the same hyperparameters for all the models. For similarity-based methods, the only hyperparameter is the max sequence length, we empirically choose a number that can fit at least $99\%$ of the samples. We respectively set it as $64$, $64$, $128$, and $128$ for intent classification, out-of-scope detection, utterance-level response selection and dialogue-level response selection. Hyperparameters for fine-tune evaluations as listed as follows:

\paragraph{Intent Classification} We fine-tune all the models for $50$ epochs with a batch size of $16$ and learning rate of $3e$-$05$. We evaluate the model on the few-shot validation set after every $10$ steps. Early stopping is applied based on the model's validation results. The max sequence length is set as $64$ and the dropout at the classification layer is set as $0.1$.

\paragraph{Utterance-level Response Selection} In this task, we set the max sequence length as $128$ and batch size as $100$. Other hyperparameters are same as those in Intent Classification. We use the original SimCLR loss \citep{simclr} to optimize the model.

\paragraph{Dialogue Action Prediction} In this task, we fine-tune all the models for $100$ epochs with a batch size of $32$ and learning rate of $5e$-$05$. We evaluate the model on the few-shot validation set after every $30$ steps. Early stopping is also applied. The max sequence length is set as $32$ since we find shorter inputs leads to much better performance for all the models. We truncate sentences from the head to keep the most recent dialogue utterances as model input.
We set the dropout at the classification layer as $0.2$.

\subsection{Datasets}
\label{subsec:evaluation_data}
\paragraph{Intent Classification} We use 4 popular publicly available datasets: \texttt{Clinc150} \citep{clinc150} with $150$ categories and $4500$ test sample, \texttt{Bank77} \citep{bank77} with $77$ categories and $3080$ test sample, \texttt{Snips} \citep{snips} with $7$ categories and $1447$ test sample, and \texttt{Hwu64} \citep{hwu64} with $64$ categories and $3853$ test sample. To apply the Clinc150 dataset in intent classification, we remove all the out-of-scope samples. We also use an internal dataset named \texttt{Appen}, whose texts are transcribed from customer recording. This dataset contains $30$ categories and $310$ test samples. There are two versions of each sentence. One is transcribed by Automatic Speech Recognition (ASR), which includes some ASR noise (e.g., transcribe errors). The other is transcribed by human annotator. We refer them respectively as \texttt{Appen-A} and \texttt{Appen-H}.  

\paragraph{Out-of-scope Detection} We use the entire \texttt{Clinc150} dataset, which contains $150$ in-scope intents and one out-of-scope intent. There are $5500$ test samples in total ($4500$ in-scope and $1000$ out-of-scope). 

\paragraph{Utterance-level Response Selection} We use the \texttt{AmazonQA} dataset \citep{amazonqa}, which contains 5334606 question-answer pairs about different products. 
Following \citet{convert}, we randomly select 300K pairs for model evaluation.

\paragraph{Dialogue-Level Response Selection} We use the \texttt{DSTC7-Ubuntu} dataset \citep{ubuntu}, which contains conversations about the Ubuntu system. Each query of this dataset comes together with one ground-truth response and $100$ candidate responses.
We combine the validation and test sets together for evaluation, which results in $6000$ evaluation samples. 

\paragraph{Dialogue Action Prediction} We use the \texttt{DSTC2} \citep{dstc2} and \texttt{GSIM} \citep{gsim} dataset processed by \citet{todbert}. These two datasets respectively contains $13$/$19$ actions and $1117$/$1039$ test samples. The average number of samples in $10$-shot and $20$-shot training is $79$ and $149$ for DSTC2; $60$ and $120$ for GSIM.

\section{Results of BERT$_\texttt{base}$}
\label{sec:bertbase_other}
In this section, we present other evaluation results on the BERT$_\texttt{base}$ model, including 1-shot and 5-shot fine-tune on intent classification (Table \ref{tb:bert_intent_ft}), 5-shot similarity-based out-of-scope detection (Table \ref{tb:bert_oos_sim_5}), and 500-shot and 1000-shot fine-tune on AmazonQA response selection (Table \ref{tb:bert_rs_amazon_ft}).

\begin{table*}[t]
	\centering
	\footnotesize
	\setlength{\tabcolsep}{2.4mm}{
	\begin{tabular}{llccccccc}\toprule
		 
		   &\textbf{BERT$_{\texttt{base}}$} & {\textbf{Clinc150}} & {\textbf{Bank77}} & {\textbf{Snips}}   & {\textbf{Hwu64}}  & {\textbf{Appen-A}} & {\textbf{Appen-H}} & {\textbf{Ave.}} \\

		\midrule
        
        \multirow{7}{*}{\rotatebox{90}{\large \texttt{1-shot}}} &
		{\textbf{SimCSE-sup}$^{\clubsuit}$  } &48.30&	35.29&	49.77&	34.09&	30.16	&38.42	&39.34 \\
		
		& {\textbf{PairSupCon}$^{\clubsuit}$ } & 50.33&	37.59&	52.53&	34.21	&\textbf{33.58}&	41.65&	41.65\\
		
		& {\textbf{DSE-dianli$^{\clubsuit}$ (ours)}} &43.07&38.02&	46.57&	30.98	&29.39	&36.77&	37.47\\
		
		\cmidrule(lr){2-9}

		& {\textbf{BERT}$^{\diamondsuit}$ } &37.01&	24.28&	52.05&	26.36&	17.87	&19.71&	29.55\\

		& {\textbf{SimCSE-unsup}$^{\diamondsuit}$}  &42.72&	33.56	&47.13	&30.19	&24.00	&32.68&	35.05\\
		
		& {\textbf{TOD-BERT}$^{\diamondsuit}$}  &39.48&	26.12&	46.13&	26.81&	13.45	&23.26	&29.21 \\
		
		& {\textbf{DSE-dropout$^{\diamondsuit}$ (ours)}}  &41.89	&30.10&	44.46&	28.06	&18.68&	20.48&	30.61 \\

		& {\textbf{DSE$^{\diamondsuit}$ (ours)} }& \textbf{55.67}	&\textbf{38.10}	&\textbf{70.67}&	\textbf{37.93}&	32.68	&\textbf{45.03}&	\textbf{46.68}\\
		
	    \hdashline
	    \\
    \multirow{7}{*}{\rotatebox{90}{\large \texttt{5-shot}}} &
     {\textbf{SimCSE-sup}$^{\clubsuit}$  }   & 85.49 &	70.16 &	88.90 &	68.86 &	61.71	 &73.29 &	74.74 \\
		
		& {\textbf{PairSupCon}$^{\clubsuit}$ }    & 85.15 &	71.00 &	86.01 &	68.12	 &\textbf{63.94}	 &73.68 &	74.65\\
		
		& {\textbf{DSE-dianli$^{\clubsuit}$ (ours)}} & 81.87	&69.33	&82.49&	64.40	&58.90&	68.52	&70.92\\
		  
		\cmidrule(lr){2-9}

		& {\textbf{BERT}$^{\diamondsuit}$ } & 84.00 &	68.51 &	85.72	 &64.79	 &55.00 &	65.52 &	70.59\\

		& {\textbf{SimCSE-unsup}$^{\diamondsuit}$}  & 83.35	 &70.08	 &86.82 &	65.61	 &59.68 &	70.03 &	72.60\\
		
		& {\textbf{TOD-BERT}$^{\diamondsuit}$}  &83.15 &	65.29 &	88.49	 &66.29	 &56.32 &	67.13 &	71.11\\
		
		& {\textbf{DSE-dropout$^{\diamondsuit}$ (ours)}} &84.14&	69.74&	87.39&	66.47	&57.74&	68.13&	72.27\\

		& {\textbf{DSE$^{\diamondsuit}$ (ours)} }&  \textbf{86.67}&	\textbf{71.52}&	\textbf{92.56}&	\textbf{70.71}&	63.71&	\textbf{75.10}&	\textbf{76.71}\\

		\bottomrule
	\end{tabular}}
	\caption{ \footnotesize  
		Results of fine-tuning all the models for 1-shot and 5-shot Intent Classification for BERT$_{\texttt{base}}$ models. $\clubsuit$: Supervised models. $\diamondsuit$: Unsupervised models
	}\label{tb:bert_intent_ft}
\end{table*}

\begin{table*}[t]
	\centering
	\footnotesize
	\setlength{\tabcolsep}{2.4mm}{
	\begin{tabular}{llccccc}\toprule
		
		 &\textbf{BERT$_{\texttt{base}}$} &  {\textbf{Accuracy}} & {\textbf{In-Accuracy}} & {\textbf{OOS-Accuracy}}   & {\textbf{OOS-Recall}} & {\textbf{Ave.}}  \\

		\midrule
		\multirow{7}{*}{\rotatebox{90}{\large \texttt{mean}$-$\texttt{std}}} &
		{\textbf{SimCSE-sup}$^{\clubsuit}$  }  & 59.65& 	69.04&	80.35&	17.40	&56.61\\
		
		&{\textbf{PairSupCon}$^{\clubsuit}$ }   &68.37&	71.03&	85.62&	56.43&	70.36 \\
		
		& {\textbf{DSE-dianli$^{\clubsuit}$ (ours)}} & 56.22	&56.92&	83.38&	53.07&	62.40\\

		\cmidrule(lr){2-7}
		
		&{\textbf{BERT}$^{\diamondsuit}$ } & 53.92&	57.25&	81.94&	38.95&	58.01\\

		&{\textbf{SimCSE-unsup}$^{\diamondsuit}$}  &55.68&	63.06&	79.58&	22.47	&55.20\\
		
		& {\textbf{DialoGPT}$^{\diamondsuit}$}  & 58.53 & 63.25 & 82.62 & 37.25 & 60.41 \\
		
		&{\textbf{TOD-BERT}$^{\diamondsuit}$}  & 53.49&	56.12&	81.75&	41.64&	58.25\\

		&{\textbf{DSE-dropout$^{\diamondsuit}$ (ours)}}  &64.21&	67.39&	83.59&	49.92	&66.28 \\
        
		&{\textbf{DSE$^{\diamondsuit}$ (ours)} } & \textbf{72.62}&	\textbf{74.77}&	\textbf{87.16}&\textbf{62.95}&	\textbf{74.38} \\

	    \hdashline
	    \\

		\multirow{7}{*}{\rotatebox{90}{\large \texttt{mean}}} &
		{\textbf{SimCSE-sup}$^{\clubsuit}$  }  &42.50&	34.95&	48.26&	76.50&	50.55\\
		
		&{\textbf{PairSupCon}$^{\clubsuit}$ }   &56.19&	47.53&	62.79&	\textbf{95.17}	&65.42 \\
		
		& {\textbf{DSE-dianli$^{\clubsuit}$ (ours)}} &46.87 &	37.75 &	59.96 &	87.92 &	58.13 \\

		\cmidrule(lr){2-7}
		
		&{\textbf{BERT}$^{\diamondsuit}$ } & 47.79&	38.40&	57.86&	90.07&	58.53\\

		&{\textbf{SimCSE-unsup}$^{\diamondsuit}$}  &45.13&	36.82&	52.01&	82.51&	54.12\\
		
		& {\textbf{DialoGPT}$^{\diamondsuit}$}  & 49.64 & 40.26 & 57.49 & 91.81 & 59.80 \\
		
		&{\textbf{TOD-BERT}$^{\diamondsuit}$}  & 47.88&	38.46&	58.43&	90.31&	58.77\\

		&{\textbf{DSE-dropout$^{\diamondsuit}$ (ours)}}  &54.45&	46.00&	61.31	&92.50&	63.56\\
        
		&{\textbf{DSE$^{\diamondsuit}$ (ours)} } & \textbf{59.20}&	\textbf{51.35}&	\textbf{64.79}&	94.53&	\textbf{67.47}\\
		\bottomrule
	\end{tabular}}
	\caption{ \footnotesize  
		Results on similarity-based 5-shot out-of-scope detection on \texttt{Clinc150} dataset. The out-of-scope threshold is respectively set as \textit{mean} (\texttt{m}) and \textit{mean-std} (\texttt{m-d}) of each sample's similarity with its closest category. See Sec. \ref{subsec:task_metric} for details. All the models use BERT$_{\texttt{base}}$ as the backbone model. $\clubsuit$: Supervised models. $\diamondsuit$: Unsupervised models.
	}\label{tb:bert_oos_sim_5}
\end{table*}

\begin{table*}[t]
	\centering
	\footnotesize
	\setlength{\tabcolsep}{2.4mm}{
	\begin{tabular}{lcccccc}\toprule
		
		\multirow{3}{*}{\textbf{BERT$_{\texttt{base}}$}} &
		\multicolumn{3}{c}{\textbf{AmazonQA 500-shot}} & 
		\multicolumn{3}{c}{\textbf{AmazonQA 1000-shot}}  \\
		\cmidrule(lr){2-4} \cmidrule(lr){5-7}
		& {\textbf{Top-1 Acc.}} & {\textbf{Top-3 Acc.}} & {\textbf{Top-10 Acc.}} & {\textbf{Top-1 Acc.}} & {\textbf{Top-3 Acc.}} & {\textbf{Top-10 Acc.}} \\

		\midrule
		{\textbf{SimCSE-sup}$^{\clubsuit}$  }  &59.02&	72.90&	84.40&	60.24&	73.91&	85.17 \\
		
		{\textbf{PairSupCon}$^{\clubsuit}$ }   & 61.24&	74.51&	85.19&	62.31&	75.36	&86.01\\
		
		\midrule
		
		{\textbf{BERT}$^{\diamondsuit}$ } & 55.63&	70.98&	83.79&	58.00&	72.67&	84.81 \\

		{\textbf{SimCSE-unsup}$^{\diamondsuit}$}  & 56.04&	70.34&82.56&	57.85&	71.95& 84.01\\
		
		{\textbf{TOD-BERT}$^{\diamondsuit}$} & 43.52	&59.29&	75.06&	46.54&	62.16&	77.15  \\
        
        {\textbf{DSE-dropout$^{\diamondsuit}$ (ours)}}  & 57.66&	72.02&	83.72&	58.66	&72.86&	84.67\\

		{\textbf{DSE$^{\diamondsuit}$ (ours)} } &\textbf{61.71}	&\textbf{75.66}&	\textbf{86.49}&	\textbf{63.02}&	\textbf{76.47}	&\textbf{87.55} \\

		\bottomrule
	\end{tabular}}
	\caption{ \footnotesize  
		Results on 500-shot and 1000-shot fine-tune evaluation on response selection on \texttt{AmazonQA} (utterance-level). All the models use BERT$_{\texttt{base}}$ as the backbone model. $\clubsuit$: Supervised models. $\diamondsuit$: Unsupervised models.
	}\label{tb:bert_rs_amazon_ft}
\end{table*}

\section{Results of Other Backbone Models}
\label{sec:other_backbone}

In this section, we present similarity-based evaluation results on other four backbone models: BERT$_{\texttt{large}}$, RoBERTa$_{\texttt{base}}$, RoBERTa$_{\texttt{large}}$, and DistilBERT$_{\texttt{base}}$. Table \ref{tb:other_intent_sim} shows the results of similarity-based intent classification and Table \ref{tb:other_rs_sim} shows the results of similarity based response selection on both utterance-level and dialogue-level. As shown in the tables, DSE leads to consistent and significant performance boost on all the backbone models.

\begin{table*}[ht]
	\centering
	\footnotesize
	\setlength{\tabcolsep}{2.4mm}{
	\begin{tabular}{llccccc}\toprule
		 
		  & & {\textbf{Clinc150}} & {\textbf{Bank77}} & {\textbf{Snips}}   & {\textbf{Hwu64}}  & {\textbf{Ave.}} \\

		\midrule
        
        \multirow{7}{*}{\rotatebox{90}{\large \texttt{1-shot}}} &
		{\textbf{BERT$_{\texttt{large}}$} } & 31.71  & 20.47 & 54.31 & 25.24 & 32.93 \\
		
		& {\textbf{BERT$_{\texttt{large}}$-DSE} } & 65.57 &  45.45 & 78.52 & 46.37 & 58.97 \\
		
		\cmidrule(lr){2-7}
		
        & {\textbf{RoBERTa$_{\texttt{base}}$} } & 34.58 &20.58 & 52.25&24.24 &32.91  \\
		
		& {\textbf{RoBERTa$_{\texttt{base}}$-DSE} } & 66.05 &45.01 &80.58 & 43.98 &58.90 \\
		
		\cmidrule(lr){2-7}
		
        & {\textbf{RoBERTa$_{\texttt{large}}$} } & 35.72 & 20.84 & 54.80& 23.57 &33.73  \\
		
		& {\textbf{RoBERTa$_{\texttt{large}}$-DSE} } & 69.23 &45.42 &73.72 &44.29  &58.16 \\
		
		\cmidrule(lr){2-7}
		
        &{\textbf{DistilBERT$_{\texttt{base}}$} } & 39.48 & 23.96 &63.00 &30.25 & 39.17 \\
		
		& {\textbf{DistilBERT$_{\texttt{base}}$-DSE} } & 60.47 & 43.52& 76.38 & 44.63 & 56.25 \\
		
	    \hdashline
	    \\
    \multirow{7}{*}{\rotatebox{90}{\large \texttt{5-shot}}} &
		{\textbf{BERT$_{\texttt{large}}$} } & 46.78 & 33.53 & 70.89 & 37.06 &47.06  \\
		
		& {\textbf{BERT$_{\texttt{large}}$-DSE} } & 80.40 & 64.49 & 89.08 & 63.00 &74.24 \\
		
		\cmidrule(lr){2-7}
		
        & {\textbf{RoBERTa$_{\texttt{base}}$} } & 53.58 &32.40 &68.90 &34.98 & 47.46 \\
		
		& {\textbf{RoBERTa$_{\texttt{base}}$-DSE} } &81.73 &64.92 & 89.67 & 62.81  &74.78 \\
		
		\cmidrule(lr){2-7}
		
        & {\textbf{RoBERTa$_{\texttt{large}}$} } & 55.43 & 33.25 &78.01 & 36.25 &50.73  \\
		
		& {\textbf{RoBERTa$_{\texttt{large}}$-DSE} } &82.52 &62.93 &86.64 &61.04  &73.28 \\
		
		\cmidrule(lr){2-7}
		
        &{\textbf{DistilBERT$_{\texttt{base}}$} } & 61.00 & 39.45&  78.90& 45.00& 56.08 \\
		
		& {\textbf{DistilBERT$_{\texttt{base}}$-DSE} } & 77.16& 60.39& 86.48& 60.81 &71.21 \\

		\bottomrule
	\end{tabular}}
	\caption{ \footnotesize  
		Results on similarity-based 1-shot and 5-shot Intent Classification with different model as the backbone. DSE leads to significant and consistent performance boost for all the models.
	}\label{tb:other_intent_sim}
\end{table*}

\begin{table*}[t]
	\centering
	\footnotesize
	\setlength{\tabcolsep}{2.4mm}{
	\begin{tabular}{lcccccc}\toprule
		
		\multirow{3}{*}{\textbf{BERT$_{\texttt{large}}$}} &
		\multicolumn{3}{c}{\textbf{AmazonQA}} & 
		\multicolumn{3}{c}{\textbf{DSTC7-Ubuntu}}  \\
		\cmidrule(lr){2-4} \cmidrule(lr){5-7}
		& {\textbf{Top-1 Acc.}} & {\textbf{Top-3 Acc.}} & {\textbf{Top-10 Acc.}} & {\textbf{Top-1 Acc.}} & {\textbf{Top-3 Acc.}} & {\textbf{Top-10 Acc.}} \\

		\midrule
		
		{\textbf{BERT$_{\texttt{large}}$} } &27.97 & 41.30 & 57.04 & 6.10 & 11.08 & 22.31 \\
		
		{\textbf{BERT$_{\texttt{large}}$-DSE} } &59.63 & 73.46 & 84.12 & 16.40 & 24.56 & 36.51 \\
		
        \\
		
        {\textbf{RoBERTa$_{\texttt{base}}$} } & 19.60& 29.67&44.70 &4.86 &9.80 &20.70 \\
		
		{\textbf{RoBERTa$_{\texttt{base}}$-DSE} } &55.69 &70.01 &81.68 &15.86 &24.25 &37.38 \\
		
        \\
		
        {\textbf{RoBERTa$_{\texttt{large}}$} } &26.68 & 37.73& 51.70&7.65 &14.50 & 26.10 \\
		
		{\textbf{RoBERTa$_{\texttt{large}}$-DSE} } & 58.13& 71.65& 82.20& 18.66&27.70 & 40.93\\
		
		\\
		
        {\textbf{DistilBERT$_{\texttt{base}}$} } &31.73 &46.47 &63.23 &6.65 &12.46 &24.98 \\
		
		{\textbf{DistilBERT$_{\texttt{base}}$-DSE} } & 56.36 & 70.11 & 81.51 &14.56 & 22.78  & 35.63 \\

		\bottomrule
	\end{tabular}}
	\caption{ \footnotesize  
		Results on 0-shot response selection on \texttt{AmazonQA} (utterance-level) and \texttt{DSTC7-Ubuntu} (dialogue-level). DSE leads to significant and consistent performance improvements on all the models.
	}\label{tb:other_rs_sim}
\end{table*}

\end{document}